\pdfoutput = 1

\documentclass[11pt,a4paper]{article}
\usepackage[hyperref]{naaclhlt2018}
\usepackage{times}
\usepackage{latexsym}

\usepackage{url}

\usepackage{xcolor}
\usepackage{caption}
\usepackage{subcaption}
\usepackage{array}
\usepackage{tabulary}
\usepackage{graphicx}

\renewcommand{\vec}[1]{\mathbf{#1}}

\makeatletter
\def\endthebibliography{%
  \def\@noitemerr{\@latex@warning{Empty `thebibliography' environment}}%
  \endlist
}
\makeatother

\aclfinalcopy 
\def\aclpaperid{1042} 

\title{Watch, Listen, and Describe: Globally and Locally Aligned Cross-Modal Attentions for Video Captioning} 

\author{Xin Wang \and Yuan-Fang Wang \and William Yang Wang \\
  University of California, Santa Barbara\\
  {\tt \{xwang,yfwang,william\}@cs.ucsb.edu}
}

\date{}

\begin{document}
\maketitle

\begin{abstract}
A major challenge for video captioning is to combine audio and visual cues. Existing multi-modal fusion methods have shown encouraging results in video understanding. However, the temporal structures of multiple modalities at different granularities are rarely explored, and how to selectively fuse the multi-modal representations at different levels of details remains uncharted. In this paper, we propose a novel hierarchically aligned cross-modal attention (HACA) framework to learn and selectively fuse both global and local temporal dynamics of different modalities. Furthermore, for the first time, we validate the superior performance of the deep audio features on the video captioning task. Finally, our HACA model significantly outperforms the previous best systems and achieves new state-of-the-art results on the widely used MSR-VTT dataset. 
\end{abstract}

\section{Introduction}

\begin{figure}
\begin{center}
\includegraphics[width=3in]{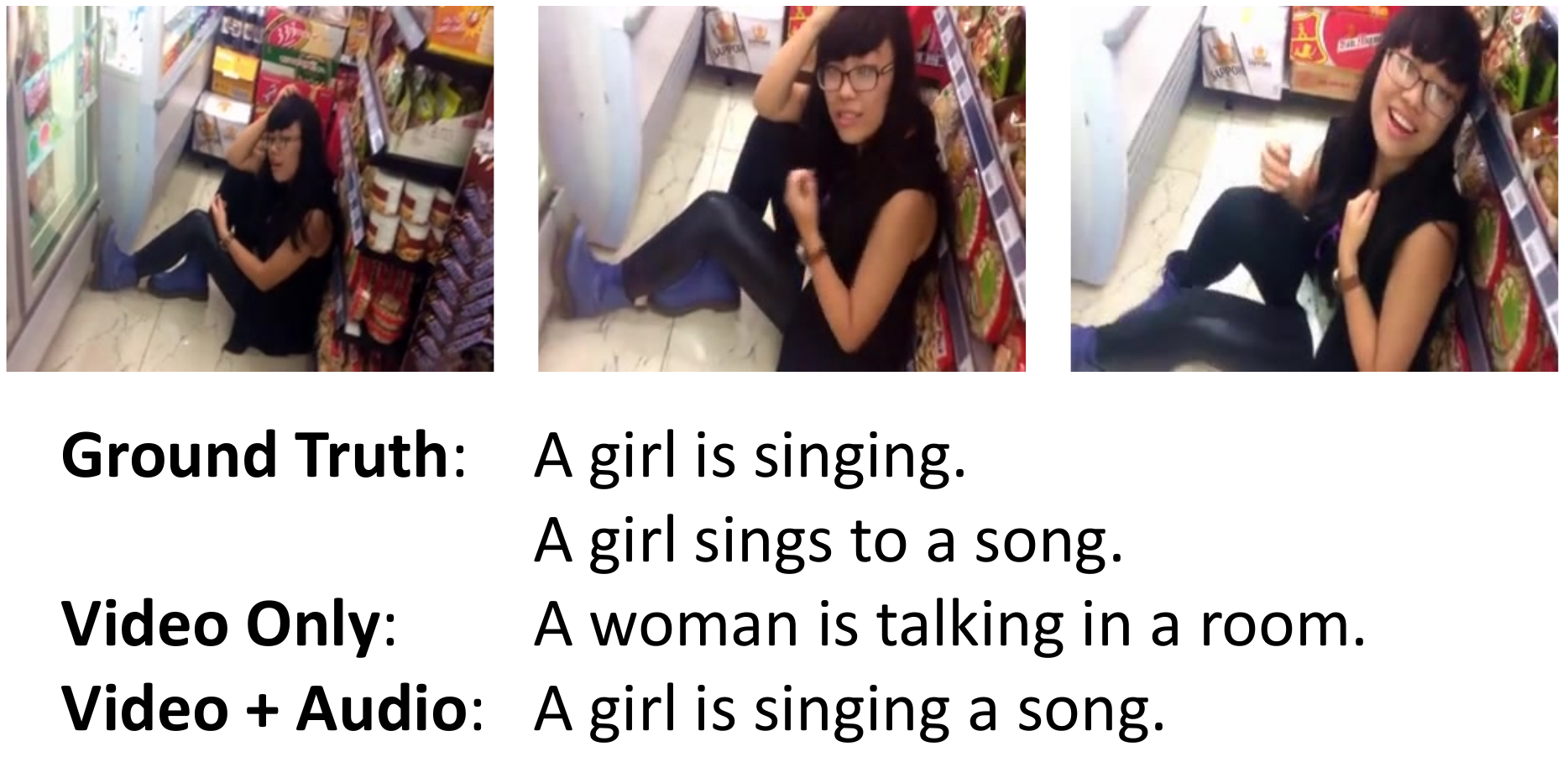}  
\end{center}
   \caption{A video captioning example.}
\label{fig:intro}
\end{figure}

Video captioning, the task of automatically generating a natural-language description of a video, is a crucial challenge in both NLP and vision communities.  In addition to visual features, audio features can also play a key role in video captioning. \autoref{fig:intro} shows an example where the caption system made a mistake analyzing only visual features. In this example, it could be very hard even for a human to correctly determine if the girl is singing or talking by only watching without listening. Thus to describe the video content accurately, a good understanding of the audio signature is a must. 

In the multi-modal fusion domain, many approaches attempted to jointly learn temporal features from multiple modalities~\cite{wu2014survey}, such as feature-level (early) fusion~\cite{ngiam2011multimodal,ramanishka2016multimodal}, decision-level (late) fusion~\cite{he2015multimodal}, model-level fusion~\cite{wu2014exploring}, and attention fusion~\cite{chen2016multi,Yang_2017_CVPR}, etc. But these techniques do not learn the cross-modal attention and thus fail to selectively attend to a certain modality when producing the descriptions.

Another issue is that little efforts have been exerted on utilizing temporal transitions of the different modalities with varying analysis granularities. The temporal structures of a video are inherently layered since the video usually contains temporally sequential activities (\textit{e.g.} a video where \textit{a person reads a book, then throws it on the table. Next, he pours a glass of milk and drinks it}). There are strong temporal dependencies among those activities. Meanwhile, to understand each of them requires understanding many action components (e.g., \textit{pouring a glass of milk} is a complicated action sequence). Therefore we hypothesize that it is beneficial to learn and align both the high-level (global) and low-level (local) temporal transitions of multiple modalities.

Moreover, prior work only employed hand-crafted audio features (\textit{e.g.} MFCC) for video captioning~\cite{ramanishka2016multimodal,xu2017learning,hori2017attention}. While deep audio features have shown superior performance on some audio processing tasks like audio event classification~\cite{hershey2017cnn}, their use in video captioning needs to be validated.  

\begin{figure*}
\begin{center}
\includegraphics[width=1\textwidth]{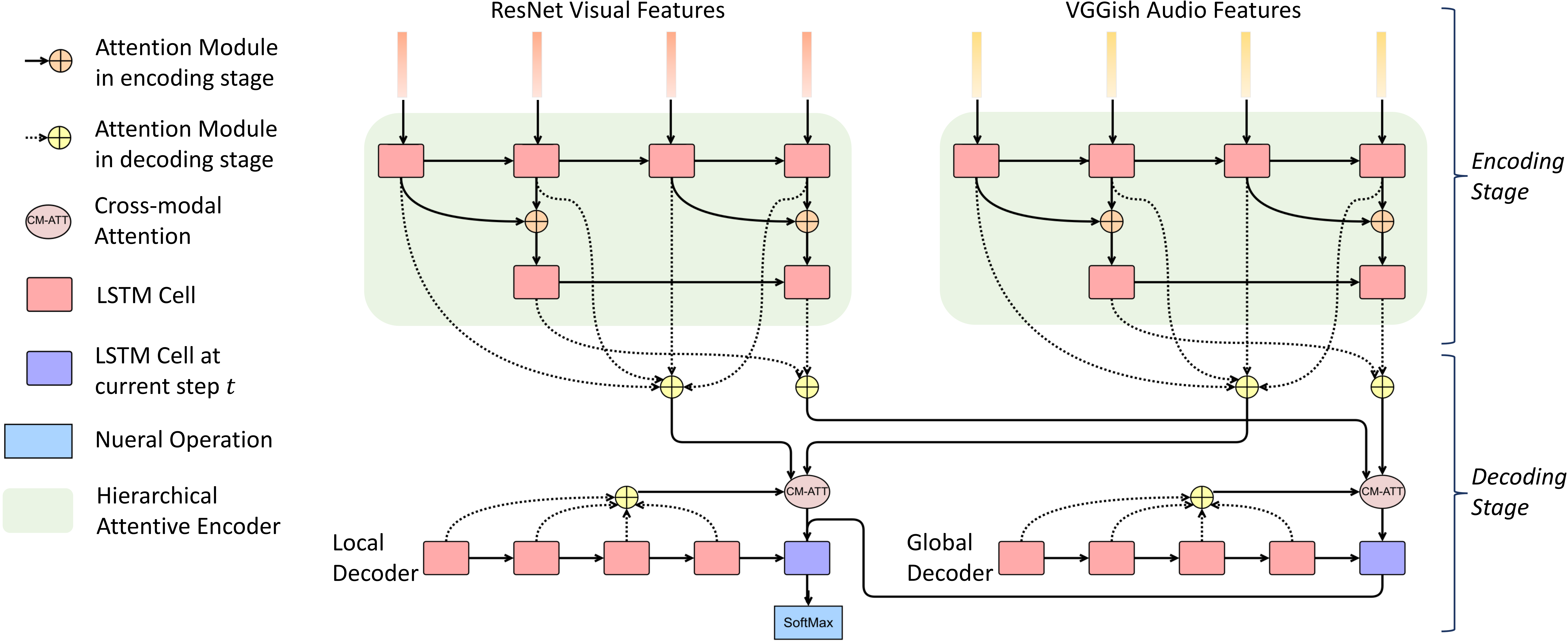}  
\end{center}
   \caption{Overview of our HACA framework. Note that in the encoding stage, for the sake of simplicity, the step size of high-level LSTM in both hierarchical attentive encoders is 2 here, but in practice usually they are set much longer. In the decoding stage, we only show the computations of the time step $t$ (the decoders have the same behavior at other time steps).}
\label{fig:framework}
\end{figure*}

In this paper, we propose a novel hierarchically aligned cross-modal attentive network (HACA) to learn and align both global and local contexts among different modalities of the video. The goal is to overcome the issues mentioned above and generate better descriptions of the input videos. Our contributions are fourfold: (1) we invent a hierarchical encoder-decoder network to adaptively learn the attentive representations of multiple modalities, including visual attention, audio attention, and decoder attention; (2) our proposed model is capable of aligning and fusing both the global and local contexts of different modalities for video understanding and sentence generation; (3) we are the first to utilize deep audio features for video captioning and empirically demonstrate its effectiveness over  hand-crafted MFCC features; and (4) we achieve the new state of the art on the MSR-VTT dataset.

Among the network architectures for video captioning~\cite{yao2015describing,venugopalan:naacl15}, sequence-to-sequence models~\cite{venugopalan2015sequence} have shown promising results. Pan et al. \citet{pan2016hierarchical} introduced a hierarchical recurrent encoder to capture the temporal visual features at different levels. \citet{yu2016video} proposed a hierarchical decoder for paragraph generation, and most recently \citet{wang2018video} invented a hierarchical reinforced framework to generate the caption phrase by phrase. But none had tried to model and align the global and local contexts of different modalities as we do. Our HACA model does only learn the representations of different modalities at different granularities, but also align and dynamically fuse them both globally and locally with hierarchically aligned cross-modal attentions. 

\section{Proposed Model}
\label{sec:model}

Our HACA model is an encoder-decoder framework comprising multiple hierarchical recurrent neural networks (see \autoref{fig:framework}). Specifically, in the encoding stage, the model has one hierarchical attentive encoder for each input modality, which learns and outputs both the local and global representations of the modality. (In this paper, visual and audio features are used as the input and hence there are two hierarchical attentive encoders as shown in \autoref{fig:framework}; it should be noted, however, that the model seamlessly extends to more than two input modalities.)

In the decoding stage, we employ two cross-modal attentive decoders: the \textit{local decoder} and the \textit{global decoder}. The global decoder attempts to align the global contexts of different modalities and learn the global cross-modal fusion context. Correspondingly, the local decoder learns a local cross-modal fusion context, combines it with the output from the global decoder, and predicts the next word. 

\subsection{Feature Extractors}
To exploit visual and audio cues, we use the pretrained convolutional neural network (CNN) models to extract deep visual features and deep audio features correspondingly. More specifically, we utilize the ResNet model for image classification~\cite{he2016deep} and the VGGish model for audio classification~\cite{hershey2017cnn}.

\subsection{Attention Mechanism}
\label{sec:att}
For a better understanding of the following sections, we first introduce the soft attention mechanism. Given a feature sequence ($\vec{x}_1, \vec{x}_2, ...,\vec{x}_n$) and a running recurrent neural network (RNN), the context vector $\vec{c}_t$ at time step $t$ is computed as a weighted sum over the sequence: 
\begin{equation}
\vec{c}_t = \sum_{k=1}^{n} \alpha_{tk} \vec{x}_k \quad ,
\end{equation}
These attention weights $\{\alpha_{tk}\}$ can be learned by the attention mechanism proposed in~\cite{bahdanau2014neural}, which gives higher weights to certain features that allow better prediction of the system's internal state. 

\subsection{Hierarchical Attentive Encoder}
Inspired by \citet{pan2016hierarchical}, the hierarchical attentive encoder consists of two LSTMs and the input to the low-level LSTM is a sequence of temporal features $\{\vec{f}_i^{e_L}\}$ and $i \in \{1,...,n\}$:
\begin{equation}
\vec{o}_i^{e_L}, \vec{h}_i^{e_L} = e_L(\vec{f}_i^{e_L}, \vec{h}_{i-1}^{e_L}) \quad,
\end{equation}
where $e_L$ is the low-level encoder LSTM, whose output and hidden state at step $i$ are $\vec{o}_i^{e_L}$ and $\vec{h}_i^{e_L}$ respectively. As shown in \autoref{fig:framework}, different from a stacked two-layer LSTM, the high-level LSTM here operates at a lower temporal resolution and runs one step every $s$ time steps. Thus it learns the temporal transitions of the segmented feature chunks of size $s$. 
Furthermore, an attention mechanism is employed between the connection of these two LSTMs. It learns the context vector of the low-level LSTM's outputs of the current feature chunk, which is then taken as the input to the high-level LSTM at step $j$. In formula,
\begin{equation}
\vec{f}_j^{e_H} = \sum_{k=s(j-1)+1}^{sj} \alpha_{jk} \vec{o}_k^{e_L} \quad,
\end{equation}
\begin{equation}
\vec{o}_j^{e_H}, \vec{h}_j^{e_H} = e_H(\vec{f}_j^{e_H}, \vec{h}_{j-1}^{e_H}) \quad,
\end{equation}
where $e_H$ denotes the high-level LSTM whose output and hidden state at $j$ are $\vec{o}_j^{e_H}$ and $\vec{h}_j^{e_H}$.

Since we are utilizing both the visual and audio features, there are two hierarchical attentive encoders ($v$ for visual features and $a$ for audio features). Hence four sets of representations are learned in the encoding stage: high-level and low-level visual feature sequences ($\{\vec{o}_{j}^{v_H}\}$ and $\{\vec{o}_i^{v_L}\}$), and high-level and low-level audio feature sequences ($\{\vec{o}_j^{a_H}\}$ and $\{\vec{o}_i^{a_L}\}$). 

\subsection{Globally and Locally Aligned Cross-modal Attentive Decoder}

In the decoding stage, the representations of different modalities at the same granularity are aligned separately with individual attentive decoders. That is, one decoder is employed to align those high-level features and learn a high-level (global) cross-modal embedding. Since the high-level features are the temporal transitions of larger chunks and focus on long-range contexts, we call the corresponding decoder as \textit{global decoder} ($d_G$). Similarly, the companion \textit{local decoder} ($d_L$) is used to align the low-level (local) features that attend to fine-grained and local dynamics.

At each time step $t$, the attentive decoders learn the corresponding visual and audio contexts using the attention mechanism (see \autoref{fig:framework}). In addition, our attentive decoders also uncover the attention over their own previous hidden states and learn aligned decoder contexts $\vec{c}_t^{d_L}$ and $\vec{c}_t^{d_G}$:
\begin{equation}
\vec{c}_t^{d_L} = \sum_{k=1}^{t-1} \alpha_{tk}^{d_L} \vec{h}_k^{d_L}, \quad \vec{c}_t^{d_G} = \sum_{k=1}^{t-1} \alpha_{tk}^{d_G} \vec{h}_k^{d_G} .
\end{equation}
\citet{paulus2017deep} also show that decoder attention can mitigate the phrase repetition issue. 

Each decoder is equipped with a \textit{cross-modal attention}, which learns the attention over contexts of different modalities. The cross-modal attention module selectively attends to different modalities and outputs a fusion context $\vec{c}_t^{f}$:
\begin{equation}
\vec{c}_t^{f} = \tanh(\beta_{tv} \vec{W}_v \vec{c}_t^v + \beta_{ta} \vec{W}_a \vec{c}_t^a + \beta_{td} \vec{W}_d \vec{c}_t^d + b),
\end{equation}
where $\vec{c}_t^v$, $\vec{c}_t^a$, and $\vec{c}_t^d$ are visual, audio and decoder contexts at step $t$ respectively; $\vec{W}_v$, $\vec{W}_a$ and $\vec{W}_d$ are learnable matrices; $\beta_{tv}$, $\beta_{ta}$ and $\beta_{td}$ can be learned in a similar manner of the attention mechanism in Section~\ref{sec:att}.

The global decoder $d_G$ directly takes as the input the concatenation of the global fusion context $\vec{c}_t^{f_G}$ and the word embedding of the generated word $w_{t-1}$ at previous time step:
\begin{equation}
\vec{o}_t^{d_G}, \vec{h}_t^{d_G} = d_G([\vec{c}_t^{f_G}, emb(w_{t-1})], \vec{h}_{t-1}^{d_G}). 
\end{equation}
The global decoder's output $\vec{o}_t^{d_G}$ is a latent embedding which represents the aligned global temporal transitions of multiple modalities.  Differently, the local decoder $d_L$ receives the latent embedding $\vec{o}_t^{d_G}$, mixes it with the local fusion context $c_t^{f_L}$ , and then learns a uniform representation $\vec{o}_t^{d_L}$ to predict the next word. In formula,
\begin{equation}
\vec{o}_t^{d_L}, \vec{h}_t^{d_L} = d_L([\vec{c}_t^{f_L}, emb(w_{t-1}),\vec{o}_t^{d_G}], \vec{h}_{t-1}^{d_L}).
\end{equation}

\subsection{Cross-Entropy Loss Function}
The probability distribution of the next word is 
\begin{equation}
p(w_t|w_{1:t-1}) = softmax(\vec{W}_p [\vec{o}_t^{d_L}]),
\end{equation}
where $\vec{W}_p$ is the projection matrix. $w_{1:t-1}$ is the generated word sequence before step $t$. 
$\theta$ be the model parameters and $w_{1:T}^*$ be the ground-truth word sequence, then the cross entropy loss
\begin{equation}
\mathcal{L}(\theta) = - \sum_{t=1}^T \log p(w_t^* | w_{1:t-1}^*, \theta).
\end{equation}

\section{Experimental Setup}
\paragraph{Dataset and Preprocessing}
We evaluate our model on the MSR-VTT dataset~\cite{xu2016msr}, which contains 10,000 videos clips (6,513 for training, 497 for validation, and the remaining 2,990 for testing). Each video contains 20 human annotated reference captions collected by Amazon Mechanical Turk. 
To extract the visual features, the pretrained ResNet model~\cite{he2016deep} is used on the video frames which are sampled at $3fps$. For the audio features, we process the raw WAV files using the pretrained VGGish model as suggested in~\citet{hershey2017cnn}\footnote{{\small\url{ https://github.com/tensorflow/models/tree/master/research/audioset}}}.

\paragraph{Evaluation Metrics}
We adopt four diverse automatic evaluation metrics: BLEU, METEOR, ROUGE-L, and CIDEr-D, which are computed using the standard evaluation code from MS-COCO server~\cite{chen2015microsoft}.

\paragraph{Training Details}
All the hyperparameters are tuned on the validation set. The maximum number of frames is 50, and the maximum number of audio segments is 20. For the visual hierarchical attentive encoders (HAE), the low-level encoder is a bidirectional LSTM with hidden dim 512 (128 for the audio HAE), and the high-level encoder is an LSTM with hidden dim 256 (64 for the audio HAE), whose chunk size $s$ is 10 (4 for the audio HAE). The global decoder is an LSTM with hidden dim 256 and the local decoder is an LSTM with hidden dim 1024. The maximum step size of the decoders is 16. We use word embedding of size 512. Moreover, we adopt Dropout~\cite{srivastava2014dropout} with a value 0.5 for regularization. The gradients are clipped into the range [-10, 10]. We initialize all the parameters with a uniform distribution in the range [-0.08, 0.08]. Adadelta optimizer~\cite{zeiler2012adadelta} is used with batch size 64. The learning rate is initially set as 1 and then reduced by a factor 0.5 when the current CIDEr score does not surpass the previous best for 4 epochs. The maximum number of epochs is set as 50, and the training data is shuffled at each epoch. Schedule sampling~\cite{bengio2015scheduled} is employed to train the models. Beam search of size 5 is used during the test time inference.

\section{Results}

\begin{table} 
\small
\renewcommand{\arraystretch}{1.05}
\begin{center}
  \begin{tabular}{ l @{\hspace{0cm}} c @{\hspace{0.2cm}} c @{\hspace{0.2cm}} c @{\hspace{0.2cm}} c }
  
   Models & BLEU-4 & METEOR & ROUGE-L & CIDEr \\
   \hline
   \multicolumn{5}{c}{Top-3 Results from MSR-VTT Challenge 2017}\\
   \hline
   v2t\_navigator     & 40.8 & 28.2 & 60.9 & 44.8  \\
   Aalto             & 39.8 & 26.9 & 59.8 & 45.7  \\
   VideoLAB         & 39.1 & 27.7 & 60.6 & 44.1    \\
   \hline
   \multicolumn{5}{c}{State Of The Arts}\\
   \hline
   CIDEnt-RL        & 40.5 & 28.4 & 61.4 & \textbf{51.7}    \\
   Dense-Cap        & \textbf{41.4} & 28.3 & 61.1 & 48.9    \\
   HRL                & 41.3 & \textbf{28.7} & \textbf{61.7} & 48.0    \\
   \hline
   \multicolumn{5}{c}{Our Models}\\
   \hline
   ATT(v)            & 39.6 & 27.4 & 59.7 & 45.8 \\
   CM-ATT(va)         & 41.7 & 28.6 & 61.2 & 48.2 \\ 
   CM-ATT(vad)         & 41.9 & 29.1 & 61.5 & 48.0 \\ 
   HACA(w/o align)    & 42.8 & 29.0 & \textbf{61.8} & 48.9 \\
   HACA                & \textbf{43.4} & \textbf{29.5} & \textbf{61.8} & \textbf{49.7} \\ 
   
  \end{tabular}
\end{center}
\caption{Results on the MSR-VTT dataset.}
\label{table:msrvtt}
\end{table}

\subsection{Comparison with State Of The Arts}
In Table~\ref{table:msrvtt}, we first list the top-3 results from the MSR-VTT Challenge 2017: v2t\_navigator~\cite{jin2016describing}, Aalto~\cite{shetty2016frame}, and VideoLAB~\cite{ramanishka2016multimodal}. Then we compare with the state-of-the-art methods on the MSR-VTT dataset: CIDEnt-RL~\cite{pasunuru2017reinforced}, Dense-Cap~\cite{Shen_2017_CVPR}, and HRL~\cite{wang2018video}. Our HACA model significantly outperforms all the previous methods and achieved the new state of the art on BLEU-4, METEOR, and ROUGE-L scores. Especially, we improve the BLEU-4 score from 41.4 to 43.1. The CIDEr score is the second best and only lower than that of CIDEnt-RL which directly optimizes the CIDEr score during training with reinforcement learning. Note that all the results of our HACA method reported here are obtained by supervised learning only.

\subsection{Result Analysis}
We also evaluate several baselines to validate the effectiveness of the components in our HACA framework (see Our Models in \autoref{table:msrvtt}). \textit{ATT(v)} is a generic attention-based encoder-decoder model that specifically attends to the visual features only.  
\textit{CM-ATT} is a cross-modal attentive model, which contains one individual encoder for each input modality and employs a cross-modal attention module to fuse the contexts of different modalities. \textit{CM-ATT(va)} denotes the CM-ATT model consisting of visual attention and audio attention, while \textit{CM-ATT(vad)} has an additional decoder attention. 

As presented in \autoref{table:msrvtt}, our ATT(v) model achieves comparable results with the top-ranked results from MSR-VTT challenge.
Comparing between ATT(v) and CM-ATT(va), we observe a substantial improvement by exploiting the deep audio features and adding cross-modal attention. The results of CM-ATT(vad) further demonstrates that decoder attention was beneficial for video captioning. 
Note that to test the strength of the aligned attentive decoders, we provide the results of \textit{HACA(w/o align)} model, which shares almost same architecture with the HACA model, except that it only has one decoder to receive both the global and local contexts. Apparently, our HACA model obtains superior performance, which therefore proves the effectiveness of the context alignment mechanism. 

\begin{table}
\small
\renewcommand{\arraystretch}{1.05}
\begin{center}
  \begin{tabular}{ l @{\hspace{0.1cm}} c @{\hspace{0.2cm}} c @{\hspace{0.2cm}} c @{\hspace{0.2cm}} c }
  
   Features & BLEU-4 & METEOR & ROUGE-L & CIDEr \\
   \hline
   video only            & 39.6 & 27.4 & 59.7 & 45.8 \\
   video + MFCC         & 40.3 & 28.5 & 60.8 & 47.5 \\ 
   video + VGGish        & \textbf{41.7} & \textbf{28.6} & \textbf{61.2} & \textbf{48.2}
  \end{tabular}
\end{center}
\caption{Performance of the cross-modal attention model with various audio features.}
\label{table:audio}
\end{table}

\subsection{Effect of Deep Audio Features}
In order to validate the superiority of the deep audio features in video captioning, we illustrate the performance of different audio features applied in the CM-ATT model in \autoref{table:audio}. 
Evidently, the deep VGGish audio features work better than the hand-crafted MFCC audio features for the video captioning task. Besides, it also shows the importance of understanding and describing a video with the help of audio features.

\subsection{Learning Curves}
\begin{figure}
\begin{center}
\includegraphics[width=0.5\textwidth]{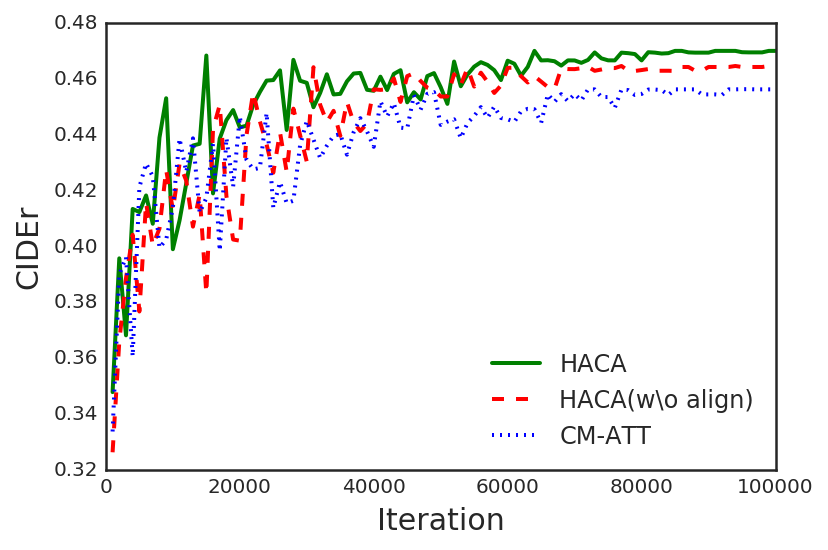}  
\end{center}
   \caption{Learning curves of the CIDEr scores on the validation set. Note that greedy decoding is used during training, while beam search is employed at test time, thus the testing scores are higher than the validation scores here.}
\label{fig:curve}
\end{figure}

For a more intuitive view of the model capacity, we plot the learning curves of the CIDEr scores on the validation set in \autoref{fig:curve}. Three models are presented: HACA, HACA(w/o align), and CM-ATT. They are trained on same input modalities and all paired with visual, audio and decoder attentions. We can observe that the HACA model performs consistently better than others and has the largest model capacity.


\section{Conclusion}
We introduce a generic architecture for video captioning which learns the aligned cross-modal attention globally and locally. It can be plugged into the existing reinforcement learning methods for video captioning to further boost the performance. Moreover, in addition to the deep visual and audio features, features from other modalities can also be incorporated into the HACA framework, such as optical flow and C3D features.

\newpage
\bibliography{naaclhlt2018}
\bibliographystyle{acl_natbib}

\end{document}